\documentclass[lettersize,journal]{IEEEtran}
\usepackage{amsmath,amsfonts}
\usepackage{algorithmic}
\usepackage{algorithm}
\usepackage{array}
\usepackage[caption=false,font=normalsize,labelfont=sf,textfont=sf]{subfig}
\usepackage{textcomp}
\usepackage{stfloats}
\usepackage{url}
\usepackage{verbatim}
\usepackage{graphicx}
\usepackage{cite}
\usepackage{booktabs}
\usepackage{multirow}
\usepackage{comment}

\begin{document}

\title{RefQSR: Reference-based Quantization for Image Super-Resolution Networks}

\author{Hongjae Lee, Jun-Sang Yoo, and Seung-Won Jung, \IEEEmembership{Senior Member, IEEE}
\thanks{\emph{(Corresponding author: Seung-Won Jung.)}}
\thanks{H. Lee, J.-S. Yoo, and S.-W. Jung are with the Department
of Electrical Engineering, Korea University, Seoul,
Korea (e-mail: jimmy9704@korea.ac.kr; junsang7777@korea.ac.kr; swjung83@korea.ac.kr).}
}

\markboth{}%
{Shell \MakeLowercase{\textit{et al.}}: A Sample Article Using IEEEtran.cls for IEEE Journals}


\maketitle

\begin{abstract}
Single image super-resolution (SISR) aims to reconstruct a high-resolution image from its low-resolution observation. Recent deep learning-based SISR models show high performance at the expense of increased computational costs, limiting their use in resource-constrained environments. As a promising solution for computationally efficient network design, network quantization has been extensively studied. However, existing quantization methods developed for SISR have yet to effectively exploit image self-similarity, which is a new direction for exploration in this study. We introduce a novel method called reference-based quantization for image super-resolution (RefQSR) that applies high-bit quantization to several representative patches and uses them as references for low-bit quantization of the rest of the patches in an image. To this end, we design dedicated patch clustering and reference-based quantization modules and integrate them into existing SISR network quantization methods. The experimental results demonstrate the effectiveness of RefQSR on various SISR networks and quantization methods. 
\end{abstract}


\section{Introduction}
\IEEEPARstart{S}{ingle} image super-resolution (SISR) is a fundamental computer vision task that aims to convert a low-resolution (LR) image into a high-resolution (HR) image. Ultra-high-definition display devices and emerging augmented/virtual reality applications have increased demand for HR content. Therefore, super-resolution (SR) is becoming crucial and continues to receive attention in the community.

With recent advances in deep neural networks (DNNs), state-of-the-art SR networks have demonstrated impressive performance in high-quality HR image reconstruction~\cite{dong2015image, lim2017enhanced, ahn2018fast, kong2022residual}. However, such high-performance SR networks have substantial computational costs, limiting their practical applicability, especially in resource-constrained environments such as smartphones, wearable devices, and embedded systems. Therefore, significant research efforts have been made to make SR networks computationally efficient. Among many other approaches, including efficient architecture design, network pruning, and knowledge distillation (KD), quantization has received the most attention due to its simplicity, strong hardware support, and increasing demands for low-precision operations~\cite{choi2018pact, wang2019haq, dong2019hawq, liu2022rethinking, peng2023mbfquant}. 

\begin{figure}[!t]
\centering
\includegraphics[width=1.0\linewidth]{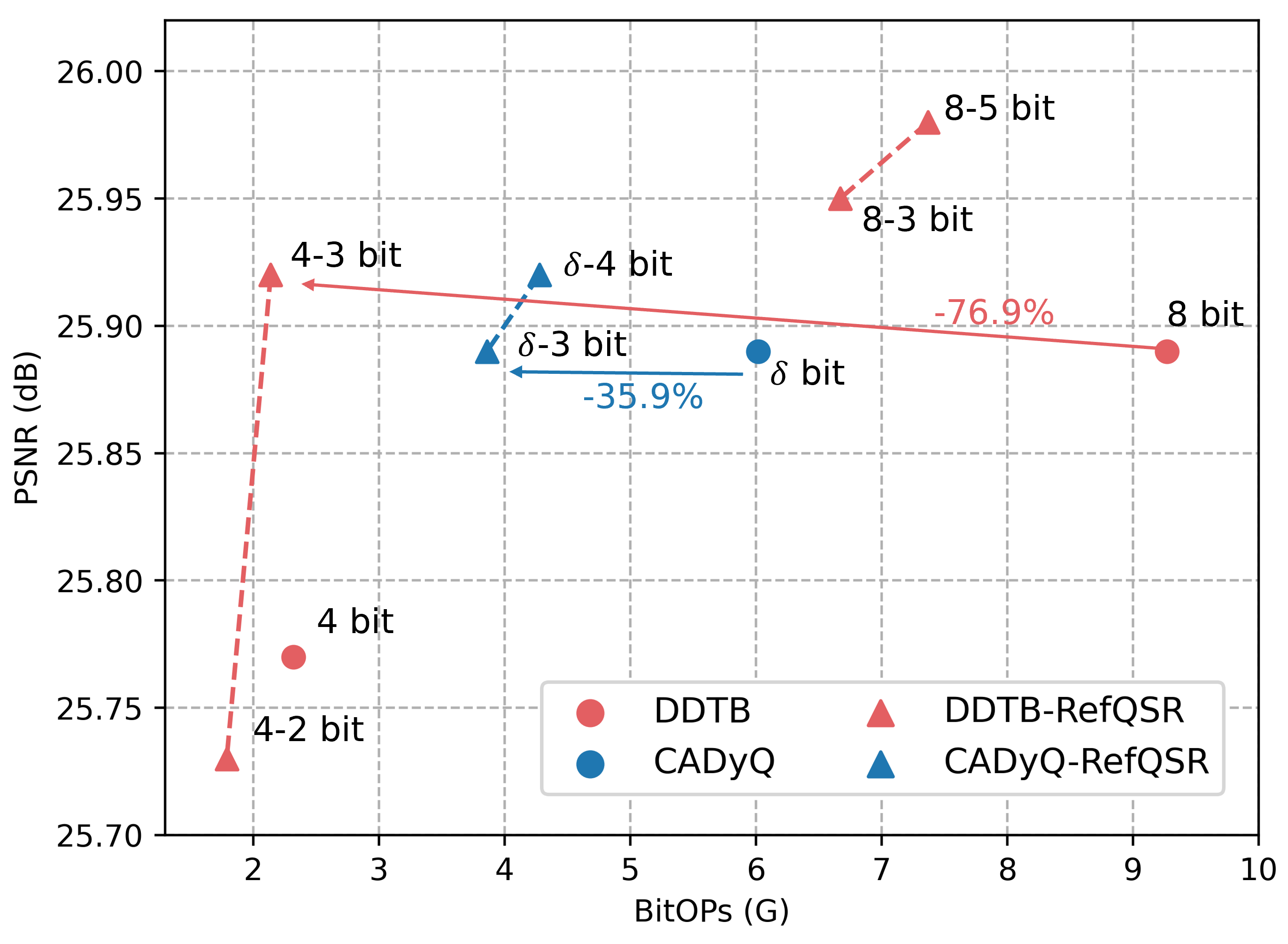}
\caption{Comparisons between the baseline SR network quantization methods and those integrated with the proposed method (denoted with -RefQSR) on $\times$4 SR of the Urban100 dataset for the SRResNet baseline. The endpoints in dashed lines correspond to several high- and low-bit combinations of RefQSR. RefQSR achieves a better trade-off between computational complexity and SR performance, \textit{e.g.}, displaying 35.9\% to 76.9\% BitOp savings over the baselines at similar PSNRs. CADyQ assigns an image-dependent dynamic bit $\delta$.}
\label{fig_1}
\end{figure}

Due to the absence of batch normalization (BN), recent SR networks exhibit apparent differences in the activation statistics compared to the networks trained on other vision tasks, such as image classification and object detection. Consequently, many quantization methods dedicated to SR have been introduced in the literature. For example, PAMS~\cite{li2020pams} dynamically adjusts the upper bound of the quantization range for each layer, and DDTB~\cite{zhong2022dynamic} further adapts both the lower and upper bounds.

Furthermore, considering that memory and computational costs increase quadratically with input size, many SR methods divide an LR input image into patches, perform patch-level SR, and merge the resultant patches to reconstruct an HR output image. Specifically, ClassSR~\cite{kong2021classsr} classifies patches into different classes based on restoration difficulties, and an SR network of appropriate complexity is applied to each class. FAD~\cite{xie2021learning} determines the complexity of the network based on image complexity using the frequency domain. 
For the quantization of patch-based SR networks, CADyQ~\cite{hong2022cadyq} and CABM~\cite{tian2023cabm} consider the characteristics of the image patches to determine the quantization precision for each layer. However, these methods do not exploit cross-patch similarities for quantization, leaving room for further improvement. In particular, numerous patches with similar patterns and structures are frequent within an image. Consequently, assigning identical bit precision to these similar patches may lead to inefficient quantization. 

We propose a novel quantization method, dubbed reference-based quantization for SR (RefQSR), to further improve the performance of existing quantization methods. Inspired by image self-similarity~\cite{huang2015single,RZSR}, RefQSR selects several representative patches and quantizes them using high-bit precision. Then, the selected patches are assigned as references and used to assist in low-bit quantization of the rest of the patches, called query patches, in an image. To mitigate quantization errors caused by low-precision processing of query patches, we introduce a reference-based error refinement (RefER) block. RefER takes features from a query patch and its associated reference patch as input and fuses them to compensate for quantization errors of the query patch. The proposed patch clustering and RefER can be integrated with existing SR network quantization methods, improving their performance in terms of computational complexity and SR performance trade-off, as demonstrated in Fig.~\ref{fig_1}.
Our contributions can be summarized as follows:
\begin{itemize}
\item {To the best of our knowledge, RefQSR is the first approach that explicitly uses image self-similarity for SR network quantization. }
\item {We propose a trainable patch clustering method to use the cluster centroids and their associated members as reference and query patches for RefQSR, respectively.}
\item {We present RefER, which can reduce quantization errors of the query patches using the features from the reference patch. RefER can be integrated into existing SR network quantization methods, improving their computational complexity and SR performance trade-off.
}

\end{itemize}
 
\begin{figure*}[!t]
\centering
\includegraphics[width=1.0\linewidth]{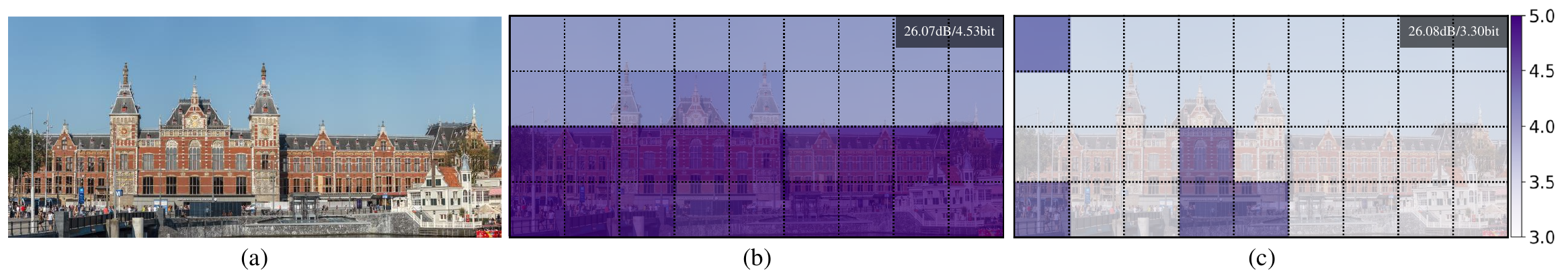}
\caption{Examples of patch-wise bit selections of (b) CADyQ and (c) CADyQ-RefQSR for the SRResNet baseline, where the darker the patches, the higher the average bit precision. Note that CADyQ allocates higher bit precision to many patches with textual details, whereas CADyQ-RefQSR utilizes image self-similarity to prevent assigning high-bit precision to similar patches.}
\label{clustering}
\end{figure*}

\section{Related Work}

\subsection{Single Image Super Resolution}
Recent advances in DNNs have motivated various DNN-based SISR methods~\cite{dong2014learning, he2016deep, zhang2018residual, lim2017enhanced}. Since the pioneering method called SRCNN~\cite{dong2014learning}, most research endeavors have been devoted to applying advanced building blocks to DNNs such as residual~\cite{he2016deep}, dense~\cite{zhang2018residual}, and Transformer blocks~\cite{vaswani2017attention}. In particular, EDSR~\cite{lim2017enhanced} discovers that BN layers degrade SR performance, and thus removing them from the SR network is common in follow-up studies. Interested readers can refer to recent articles~\cite{anwar2020deep,nasrollahi2014super,wang2020deep} for a systematic review of the literature on previous SISR networks.  

Reference-based super-resolution (RefSR)~\cite{yue2013landmark,lu2021masa,sun2022learningrefsr} is another direction towards improving SR performance, which uses additional patches or images with HR textures and detail as a reference for SR of an LR image. However, obtaining a high-quality reference is challenging in practice. To this end, several RefSR methods~\cite{lu2019georefsr, chantas2021heavy, RZSR} obtain a reference image from an LR image itself by exploiting image self-similarity. For example, RZSR~\cite{RZSR} obtains reference patches from the downsampled version of the input image using cross-scale patch matching, and Huang \textit{et al.}~\cite{huang2015single} expanded the search space by applying geometric transformations to local patches.

However, the computational demands of SR networks, including RefSR ones, limit their use in resource-constrained environments. As a result, various methods have been proposed to make SR networks computationally efficient~\cite{ahn2018fast, kong2022residual, gendy2023mixer}. In particular, for handling large input images, patch-wise SR has been considered an effective solution~\cite{kong2021classsr, xie2021learning, chen2022arm, wang2022adaptive}. These lightweight SR networks, however, still rely on computationally expensive floating-point operations, requiring further processing for low-precision operations.

\subsection{Network Quantization}
Many studies have explored the quantization operation to compress and accelerate DNNs. Quantization involves mapping 32-bit floating-point values of feature maps and/or weights to lower-bit values. Conventional quantization methods~\cite{choi2018pact,zhuang2018towards,jin2020adabits} typically allocate the same bit precision for all layers. However, recent developments in hardware have enabled mixed-precision operations, allowing different layers to use different precision to enhance performance and computational efficiency~\cite{apple,nvidia,bismo}. Several methods have thus been proposed to determine the bit depth of each layer by neural networks ~\cite{dong2019hawq, cai2020rethinking,yang2021fracbits,peng2023mbfquant}. However, they focus on high-level vision tasks, and the generalization to low-level tasks such as SR is challenging.

\subsection{Quantization for SR Networks}
Several methods have recently been proposed for quantization of SR networks~\cite{xin2020binarized,jiang2021training}. For example, Ayazoglu \textit{et al.}~\cite{ayazoglu2021extremely} and Du \textit{et al.}~\cite{du2021anchor} designed an efficient SR network architecture for quantization. However, these methods are only applicable to specific models and cannot be generalized to other SR networks. To address this issue, PAMS~\cite{li2020pams} learns layer-wise quantization ranges to compensate for the absence of BN in SR networks. DAQ~\cite{hong2022daq} further assigns different quantization parameters for each channel in every layer. DDTB~\cite{zhong2022dynamic} uses dual trainable bounds to accommodate asymmetric activation distributions. Moreover, CADyQ~\cite{hong2022cadyq} leverages a multi-layer perceptron (MLP) to determine the optimal bit precision for each patch and every layer based on the gradient magnitude of the given patch and the standard deviation of activations, achieving more efficient quantization even below 8-bit precision. CABM~\cite{tian2023cabm} takes a step further by converting the MLP into a lookup table, reducing computational overhead. However, CADyQ and CABM do not consider cross-patch similarities during quantization, leaving room for further enhancements, as shown in Fig.~\ref{clustering}. Consequently, we propose a novel quantization framework, called RefQSR, that utilizes image self-similarity to avoid redundant quantization in similar patches.

\begin{figure*}[!t]
\centering
\includegraphics[width=\linewidth]{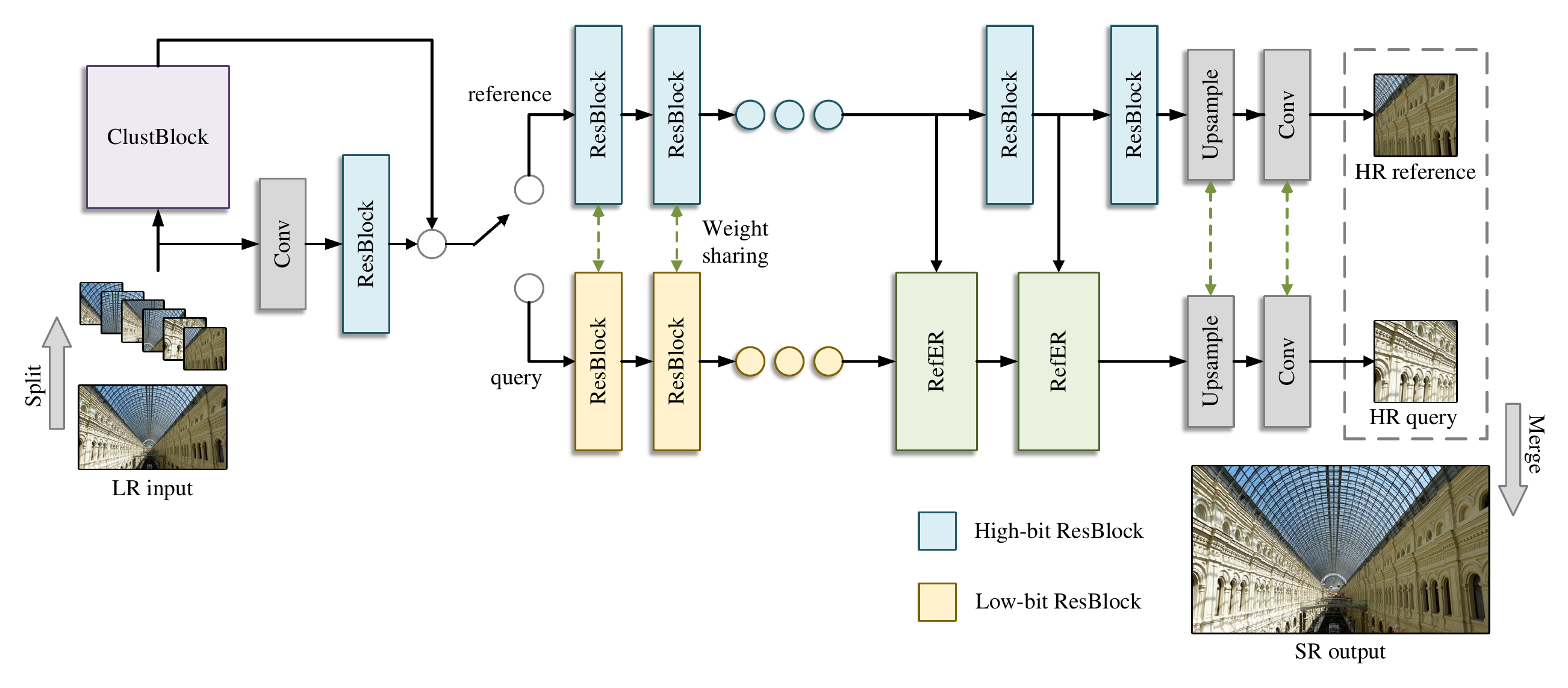}
\caption{Overview of the proposed RefQSR framework. RefQSR divides the input image into patches and clusters them as reference and query patches via ClustBlock. The ResBlocks are processed with high-bit and low-bit quantization for reference and query patches, respectively. The RefER blocks are used to restore the quantization error of the query patches using the features from the reference patch. Finally, the output reference and query patches are merged to reconstruct the final HR image.}
\label{overview}
\end{figure*}

\section{Proposed Method}

\subsection{Preliminaries}
Since our method can be integrated with various quantization methods~\cite{li2020pams,zhong2022dynamic,hong2022cadyq,tian2023cabm}, we first introduce the background of network quantization. Given weights or activations, denoted as $\pmb{X}$, the quantized output $\pmb{X}_{q_b}$ can be obtained as follows:
\begin{equation}
\label{eqn: quantization}
\pmb{X}_{{q}_{b}} = Q_{b}\left(\pmb{X} \right) = round\left(\frac{clip(\pmb{X},\alpha ,\beta )}{s(b)}\right)\cdot s(b),
\end{equation}
where $clip(\pmb{X},\alpha ,\beta ) = \min(\max(\pmb{X},\alpha),\beta)$ is a function that clips the input $\pmb{X}$ to the range $[\alpha, \beta]$, and the function $s(b) = \frac{\beta - \alpha }{2^{b}-1}$ maps floating-point numbers to their corresponding quantization precision $b$. In high-level vision tasks, it is common to set $\alpha=\min\left(|\pmb{X}|\right)$ and $\beta=\max\left(|\pmb{X}|\right)$. However, the activations of the SR networks exhibit diverse and asymmetric distributions~\cite{lim2017enhanced,zhang2018residual,ahn2018fast}, requiring $\alpha$ and $\beta$ to be learnable parameters.

\subsection{Motivation}
Most SR network quantization methods use a fixed quantization precision $b$ for all layers~\cite{li2020pams,zhong2022dynamic}. Some exceptional works~\cite{hong2022cadyq,tian2023cabm} adjust $b$ for each layer according to the signal characteristics of the patch. However, our observations reveal the drawback of determining the quantization precision for each patch independently. Specifically, as depicted in Fig.~\ref{clustering}, many patches with similar patterns and structures frequently occur within an image; thus, assigning the same bit precision to those similar patches can result in inefficient quantization. To exploit such image self-similarity, the proposed RefQSR assigns a high value of $b$ to a few representative patches, called reference patches, and a low value of $b$ to the rest of the patches, called query patches.

\subsection{Reference-based Quantization for SR } \label{sec:prop-refqsr}
\subsubsection{Framework overview}
The proposed RefQSR framework consists of a patch clustering block called ClustBlock, ResBlocks, convolutional layers, and RefER blocks, as shown in Fig.~\ref{overview}. RefQSR divides the input image into patches, which are then classified as reference patches and query patches via ClustBlock. Note that ClustBlock is used only in the inference stage, and the reference and query patches are obtained differently in the training stage, which will be detailed in Section~\ref{sec:Experiment}. Following previous work~\cite{li2020pams,zhong2022dynamic,hong2022cadyq,tian2023cabm}, we do not quantize the low-level feature extractor and the last image reconstruction layer, denoted as gray blocks in Fig.~\ref{overview}. In addition, the first ResBlock maintains a high-bit precision $b_{high}$ in both the reference and the query.

The activations of the subsequent ResBlocks are quantized using a high-bit precision $b_{high}$ for reference patches and a low-bit precision $b_{low}$ for query patches. All ResBlocks for the reference and query patches share the same weights, but the weights are quantized using $b_{high}$ for the reference patches and $b_{medium}=round\left(\frac{b_{high}+b_{low}}{2}\right)$ for the query patches. The last few ResBlocks for the query patches, \textit{e.g.}, two ResBlocks in the case of SRResNet, are replaced with the RefER blocks, which are used to leverage the reference patch to compensate for the quantization errors of the query patch. By merging super-resolved reference and query patches, RefQSR obtains a final HR image.

\begin{table}[t]
\centering
\renewcommand{\tabcolsep}{1.mm}
\caption{The structure of the feature extractor of ClustBlock}
\label{tab:clustnet}
    \begin{tabular}{cccccc}
        \toprule
        \multirow{2}{*}{Layer No.} & \multirow{2}{*}{Operator} & Kernel & \multirow{2}{*}{Stride} & \multirow{2}{*}{Padding} \\
        &&($H \times W \times C_{in} \times C_{out}$) \\
        \midrule
        1 &  QConv / ReLU & $ 3 \times 3 \times 3 \times 64$ & 3 & 0 \\
        2 &  QConv / ReLU & $ 3 \times 3 \times 64 \times 64$ & 1 & 0 \\
        3 &  QConv / ReLU & $ 3 \times 3 \times 64 \times 32$ & 1 & 0 \\
        4 &  GAP &  - &  - &  - \\
        5 &  QFC & $ 32 \times \hat{C}$ & - & - \\
        \bottomrule
    \end{tabular}
\end{table}

\begin{figure*}[!t]
\centering
\includegraphics[width=\linewidth]{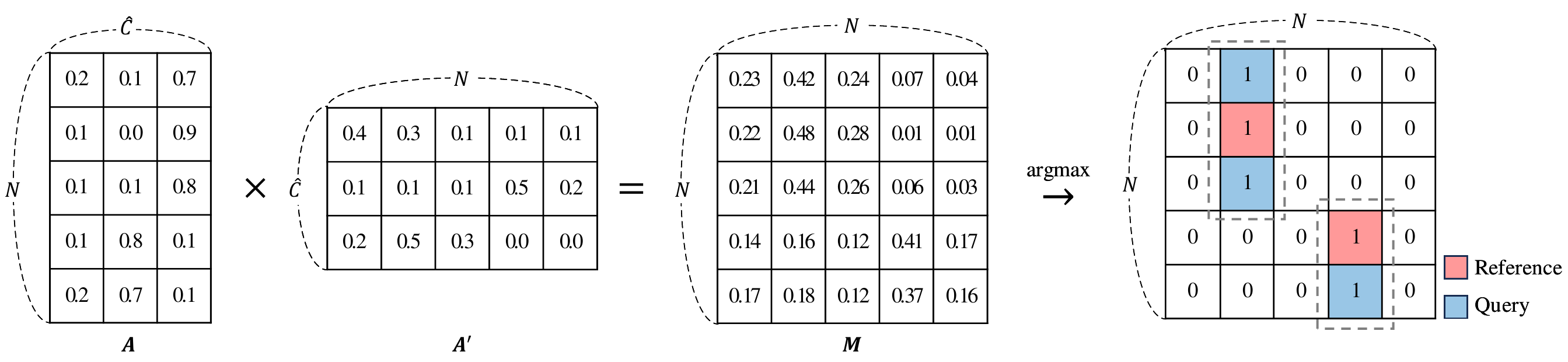}
\caption{A toy example illustrating the patch clustering procedure, where the number of patches $N$ is 5, and the number of texture classes $\hat{C}$ is 3. The five patches are grouped into two clusters according to the argmax indices, and the two patches that have the argmax indices at their positions are identified as reference patches. }
\label{clusttoy}
\end{figure*}
\subsubsection{Patch clustering}
We propose a patch clustering block called ClustBlock. In ClustBlock, a feature extractor extracts a feature vector for each patch using the network structure provided in Table~\ref{tab:clustnet}. We pre-train this feature extractor using the DTD dataset~\cite{cimpoi2014dtd} to extract textual features from patches. To ensure low complexity, we quantize the feature extractor to 8 bits using (\ref{eqn: quantization}) to build quantized convolution (QConv) and quantized fully connected (QFC) layers. The clipping ranges, $\alpha$ and $\beta$, are chosen as the minimum and maximum values for the activations and weights, respectively.

Let $\pmb{P}\in\mathbb{R}^{N\times 3 \times H \times W}$ denote a stack of $N$ patches with the size of $3\times H \times W$, collected from an input LR image. $\pmb{P}$ is fed into the feature extractor, producing a stack of $N$ feature vectors $\pmb{V}\in\mathbb{R}^{N\times \hat{C}}$, where $\hat{C}$ is the number of texture classes in the DTD dataset. To perform patch clustering using $\pmb{V}$, we first obtain a matrix $\pmb{A}\in\mathbb{R}^{N\times \hat{C}}$, defined as $\pmb{A} = \textrm{Softmax}(\pmb{V})$. Specifically, each entry $a_{ij}$ of $\pmb{A}$ is given as follows:
\begin{equation}
\label{eqn: softmax}
a_{ij} = \frac{\exp(v_{ij})}{\sum_{j}\exp(v_{ij})},
\end{equation}
where $v_{ij} = \pmb{V}(i,j)$.
Thus, each row of $\pmb{A}$ corresponds to the probability distribution of texture classes for each patch. We then obtain a matrix $\pmb{A}'\in\mathbb{R}^{\hat{C}\times N}$, defined as $\pmb{A}' = \textrm{Softmax}\left(\pmb{V}^\top\right)$, where its entry $a'_{ij}$ is given as follows:
\begin{equation}
\label{eqn: softmax2}
a'_{ij} = \frac{\exp(v_{ij})}{\sum_{i}\exp(v_{ij})}.
\end{equation}
In other words, each row of $\pmb{A}'$ corresponds to the probability distribution of patches for each texture class. Finally, let $\pmb{M}\in\mathbb{R}^{N\times N}$ = $\pmb{AA}'$. Taking the argmax function to each row of $\pmb{M}$, we can identify a reference patch for each patch. Fig.~\ref{clusttoy} illustrates the proposed clustering procedure on a toy example.

\begin{figure}[!t]
\centering
\includegraphics[width=\linewidth]{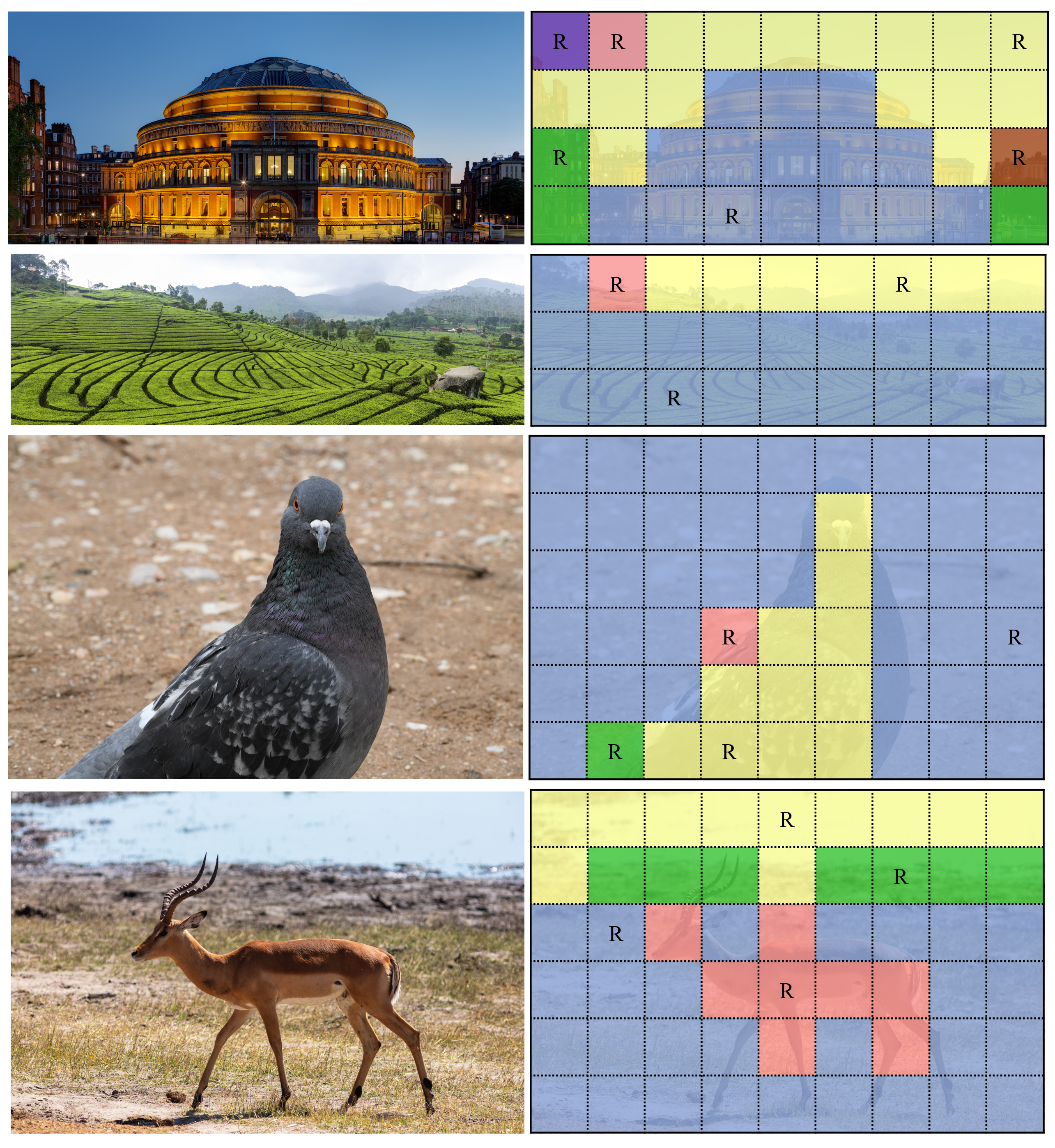}
\caption{Several images (left) and their clustering results (right): Patches with the same color belong to the same cluster. The reference patches are marked with R, where the first three samples include isolated reference patches. }
\label{fig:cluster}
\end{figure}

By the above patch clustering method, every query patch is assigned to one reference patch. However, such unconditional clustering can group less similar patches, making reference-based quantization challenging. To this end, we measure the cosine similarity between the feature vectors of the reference and query patches and do not apply reference-based quantization if the cosine similarity falls below a threshold $\tau$. In other words, query patches that are not closely matched with their reference patches are treated as isolated reference patches. Our patch clustering introduces minimal computational overhead compared to the overall SR network, as will be discussed in Section~\ref{sec:Experiment}. Fig.~\ref{fig:cluster} depicts several clustering results. 

\begin{figure}[!t]
\centering
\includegraphics[width=1.0\linewidth]{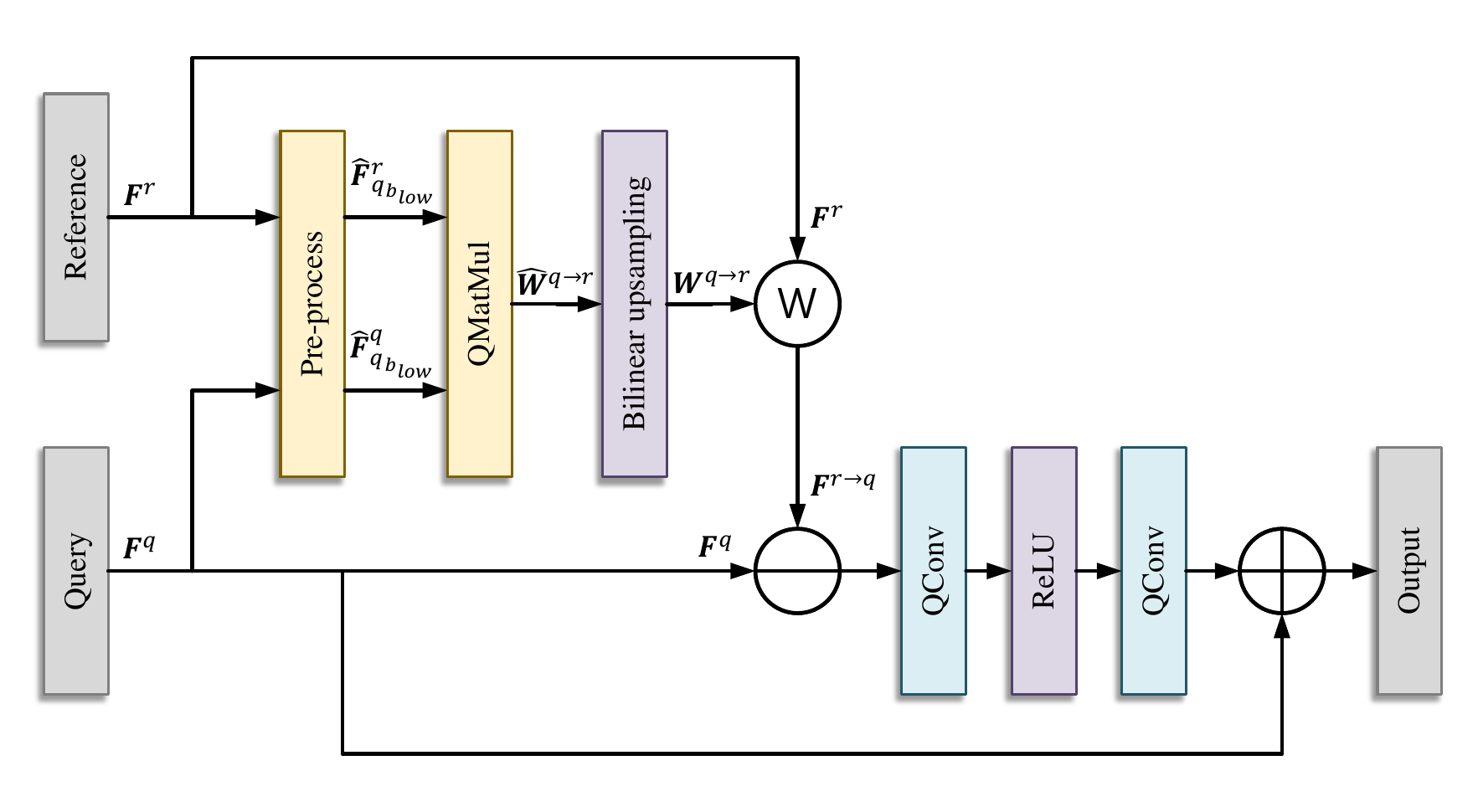}
\caption{The structure of the RefER. The reference feature is aligned with the query feature via the warping module, and the difference between the query feature and the aligned reference feature is then refined to compensate for the quantization error of the query feature. \textcircled{w} represents the backward warping operator.}
\label{refer}
\end{figure}

\subsubsection{Reference-based quantization} 
After ClustBlock, we have query patches and their associated reference patches. Our goal is to compensate for the quantization errors of the query patches by leveraging the features from the reference patches. To this end, we propose RefER, as shown in Fig.~\ref{refer}. The proposed RefER takes the features from the reference and query patches as input and produces the quantization error-reduced query feature as output. 

\begin{figure}[tb]
\centering
\includegraphics[width=1.0\linewidth]{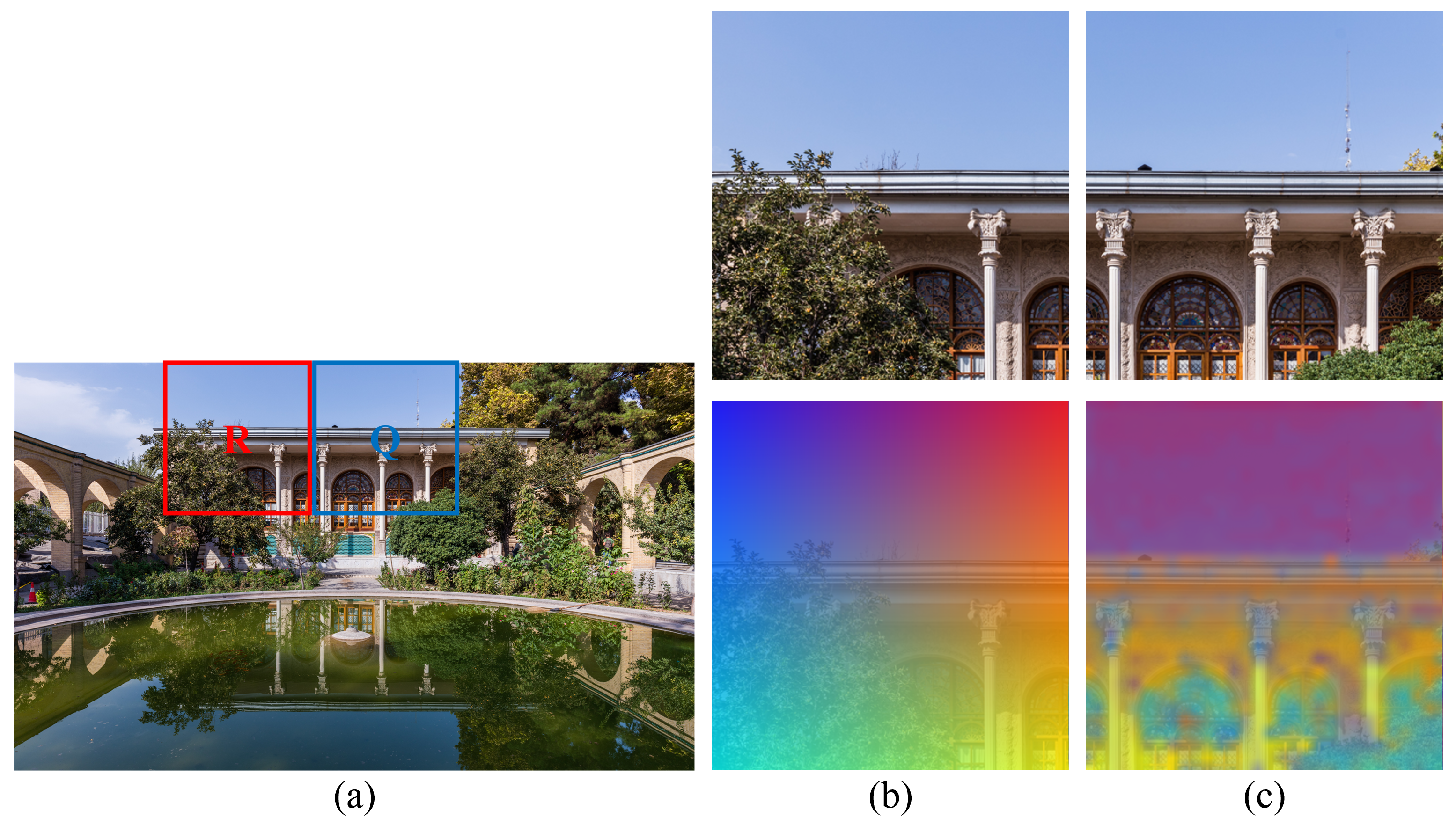}
\caption{Visualization of the correspondences between reference and query patches: (a) Input image with the reference and query patches denoted as R and Q, respectively, (b) the reference patch and the overlaid color map, (c) the query patch and the overlaid color map that is backward warped from the reference patch using the estimated flow.}
\label{fig:refer_viz}
\end{figure}

Let $\pmb{F}^r$ and $\pmb{F}^q$ denote the reference feature and the query feature, respectively. To match their bit precision to $b_{low}$ and reduce computational complexity, we first pre-process them as follows:
\begin{equation}
\label{eqn: preprocess}
\hat{\pmb{F}}^*_{q_{b_{low}}} = Q_{b_{low}}\left(\phi\left(\frac{{\pmb{F}}^*}{\left\| {\pmb{F}}^*\right\|} \right)\right)    
\end{equation}
where $* \in \left\{ {r,q} \right\}$, $\left\|  \cdot  \right\|$ measures the L2 norm, and $\phi(\cdot)$ is an adaptive average pooling with a 3×3 kernel. Since the reference and query patches are not aligned in pixels, we warp the reference feature $\pmb{F}^r$ to the query feature $\pmb{F}^q$. To this end, we compute a widely used 4D cost volume~\cite{lu2021masa, lee2020sfnet, lee2022gps}. Specifically, let $\pmb{E}$ denote the 4D cost volume between $\hat{\pmb{F}}^r_{q_{b_{low}}}$ and $\hat{\pmb{F}}^q_{q_{b_{low}}}$, defined as follows:
\begin{equation}
\label{eqn: cost volume}
\pmb{E}(\pmb{i},\pmb{j})  = \hat{\pmb{F}}^q_{q_{b_{low}}} \left(\pmb{i}\right)^\top \hat{\pmb{F}}^r_{q_{b_{low}}} (\pmb{j}),
\end{equation}
where $\pmb{i}$ and $\pmb{j}$ represent 2D coordinates, and $\hat{\pmb{F}}^q_{q_{b_{low}}} (\pmb{i})$ and $\hat{\pmb{F}}^r_{q_{b_{low}}} (\pmb{j})$ are the $C$-dimensional feature vectors at $\pmb{i}$ and $\pmb{j}$, respectively. Instead of using the standard argmax function to obtain the warping flow, we obtain $\hat{\pmb{E}}$ by applying soft-argmax to ${\pmb{E}}$, allowing backpropagation through the warping operation during training as follows:
 \begin{equation}
\label{eqn: flow softmax}
\hat{\pmb{E}}(\pmb{i},\pmb{j}) = \frac{\exp(\alpha \cdot \pmb{E}( \pmb{i},\pmb{j} ))}{{\sum\limits_{{{\pmb{k}}} \in \pmb{{\Omega}}} \exp ( \alpha  \cdot \pmb{E}( \pmb{i},{\pmb{k}} ) ) }},
\end{equation}
where $\pmb{{\Omega}}$ is the set of 2D positions, and $\alpha$ is a softmax temperature parameter, which is set to 1000 in our experiments. In the inference stage, argmax is used. The backward warping flow from the query to the reference, denoted as $\hat{\pmb{W}}^{q\to r}$, is obtained as 
\begin{equation}
\label{eqn: flow}
{\hat{\pmb{W}}^{q \to r}}\left( {\pmb{i}} \right) = \sum\limits_{{\pmb{j}} \in {\pmb{\Omega}}} {\hat{\pmb{E}}\left( {{\pmb{i}},{\pmb{j}}} \right) \cdot {\pmb{j}}}.
\end{equation}
We then apply bilinear interpolation to $\hat{\pmb{W}}^{q\to r}$, resulting in the original-sized flow $\pmb{W}^{q \to r}$. Fig.~\ref{fig:refer_viz} visualizes $\pmb{W}^{q \to r}$, indicating that our RefER can effectively identify the correspondences between the reference and query patches. Finally, we obtain the warped feature, denoted as $\pmb{F}^{r\to q}$, by applying backward warping using $\pmb{W}^{q\to r}$ and $\pmb{F}^r$. 

The RefER block obtains the difference between $\pmb{F}^q$ and $\pmb{F}^{r\to q}$, which is an essential source of information to recover the quantization error of the query feature since $\pmb{F}^{r\to q}$ is originated from the high-precision layers. Using a residual connection from $\pmb{F}^q$, RefER can concentrate on refining the quantization error of the query feature. The convolutional layers in RefER, denoted as QConv in Fig.~\ref{refer}, are processed with $b_{high}$ to better extract the signal for refinement.

\subsection{Loss functions}
In previous SR quantization methods~\cite{hong2022cadyq,tian2023cabm,li2020pams,zhong2022dynamic}, the objective function for training an SR network is defined as follows:
\begin{equation}
\label{eqn: loss_prev}
L = \lambda_1L_1 + \lambda_{kd}L_{kd},
\end{equation}
where $\lambda_1$ and $\lambda_{kd}$ are the weighting parameters to control the balance between the two loss terms. $L_1$ represents the standard L1 loss between the ground-truth and reconstructed images, while $L_{kd}$ denotes the KD loss between the full precision network and the quantized network.

In RefQSR, however, we have the outputs from both low-precision and high-precision paths, as shown in Fig.~\ref{overview}. Consequently, we modify the objective function as follows:
\begin{equation}
\label{eqn: loss_final}
L = \lambda_1(L_1^r + L_1^q) + \lambda_{kd}(L_{kd}^r + L_{kd}^q),
\end{equation}
where $L_1^r$ and $L_1^q$ are the L1 losses for the reference and query patches, respectively, and $L_{kd}^r$ and $L_{kd}^q$ are the KD losses from the full precision network for the reference and query patches, respectively.

\section{Experimental Results}\label{sec:Experiment}
\subsection{Implementation Details}
\subsubsection{Baseline quantization methods and SR networks}
For performance verification, we applied the proposed RefQSR to existing SR network quantization methods, including fixed-precision ones: PAMS~\cite{li2020pams} and DDTB~\cite{zhong2022dynamic}, and mixed-precision ones: CADyQ~\cite{hong2022cadyq} and CABM~\cite{tian2023cabm}. The quantized models using RefQSR are denoted by appending ``-RefQSR'' to their names. In addition, to demonstrate the effectiveness of RefQSR across various SR networks, we experimented with three representative SR networks: SRResNet~\cite{ledig2017photo}, CARN~\cite{ahn2018fast}, and ELAN~\cite{zhang2022elan}. For SRResNet-RefQSR and CARN-RefQSR, ResBlock was replaced with RefER. For ELAN, we used the ELAN-light version, which incorporates 24 efficient long-range attention blocks (ELABs). For ELAN-RefQSR, the local feature extractor in ELAB was replaced by RefER. Since quantization of attention scores after softmax leads to significant performance drops in Transformer-based architectures~\cite{pandey2023softmaxbias,li2023qdiff}, full-precision operations were used in the GMSA layer of ELAN. When applying RefQSR to SRResNet, CARN, and ELAN, two, one, and two RefER blocks were used, respectively. All baseline models were implemented using the official code and default settings.

\subsubsection{Training details}
We trained all models using the DIV2K training dataset~\cite{Agustsson_2017_CVPR_Workshops}. During training, we collected reference and query patches from the training dataset, as described in Algorithm~\ref{alg:mini-batch}. Specifically, since reliable pairs of reference and query patches have to be used to train SR models with RefQSR, we randomly cropped two patches from a single image and used them as reference and query patches if their ClustBlock feature distance in terms of cosine similarity (see Section~\ref{sec:prop-refqsr}) is above $\tau$, chosen as 0.5. We collected 16 pairs of reference and query patches with a size of 48$\times$48 that satisfy the above condition to construct a mini-batch for training. We used the ADAM optimizer with $\beta_1 = 0.9$, $\beta_2 = 0.999$, and $\epsilon = 10^{-8}$ and trained SRResNet, CARN, and ELAN for 400, 800, and 400 epochs, respectively. The learning rate was initialized to $10^{-4}$ and halved every 150, 300, and 150 epochs for SRResNet, CARN, and ELAN, respectively. The lookup table for CABM-RefQSR was created based on CADyQ-RefQSR. The other training strategy and hyperparameters were the same as the baseline quantization methods. Before training RefQSR, ClustBlock's feature extractor was pre-trained using the DTD dataset~\cite{cimpoi2014dtd} and quantized with 8bit using uniform quantization.

\begin{figure}[!t]
\centering
\includegraphics[width=1.0\linewidth]{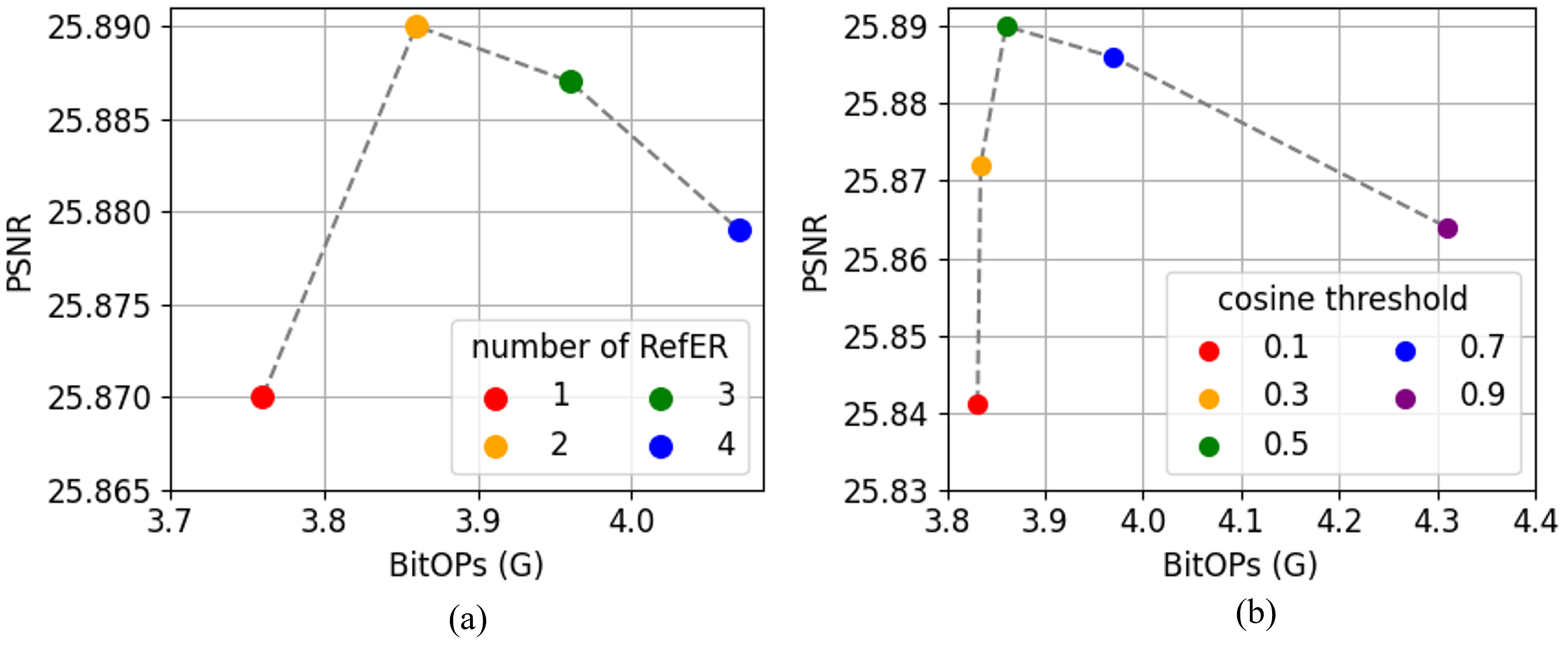}
\caption{Results on SRResNet-CADyQ using different hyperparameter settings of RefQSR: (a) the number of RefER blocks, (b) cosine similarity threshold $\tau$.}
\label{fig:hyperparameter}
\end{figure}

\subsubsection{Evaluation settings}

\begin{algorithm}
\caption{Mini-batch generation for RefQSR training} \label{alg:qdiff}
\textbf{Input:} Training image sets $\{\pmb{I}_{LR},\pmb{I}_{HR}\}$, pretrained feature extractor of ClustBlock $C_{fe}$, cosine similarity threshold $\tau$, and mini-batch size $N$. \\
\textbf{Output:} Query $\{\pmb{P}^q_{LR}, \pmb{P}^q_{HR}\}$ and reference $\{\pmb{P}^r_{LR}, \pmb{P}^r_{HR}\}$ training patches.
\begin{algorithmic}[1] 
\STATE $\{\pmb{P}^q_{LR}, \pmb{P}^q_{HR}\} \leftarrow \{\emptyset,\emptyset\}$; $\{\pmb{P}^r_{LR}, \pmb{P}^r_{HR}\} \leftarrow \{\emptyset,\emptyset\}$ 
\FOR {$n = 1, \ldots, N$}
\STATE Sample a pair of images $\{\pmb{i}_{LR},\pmb{i}_{HR}\}$ from $\{\pmb{I}_{LR},\pmb{I}_{HR}\}$
\STATE $\{\pmb{p}^q_{LR},\pmb{p}^q_{HR}\} \leftarrow$ Random crop$\left(\{\pmb{i}_{LR},\pmb{i}_{HR}\}\right)$
\STATE $\pmb{V}^q \leftarrow C_{fe}\left(\pmb{p}^q_{LR}\right)$ 
    \REPEAT
        \STATE $\{\pmb{p}^r_{LR},\pmb{p}^r_{HR}\} \leftarrow$ Random crop$\left(\{\pmb{i}_{LR},\pmb{i}_{HR}\}\right)$
        \STATE $\pmb{V}^r \leftarrow C_{fe}\left(\pmb{p}^r_{LR}\right)$
        \STATE Compute cosine similarity between query and reference patches:
        \begin{align}
          \hspace{-.15\linewidth} s_{qr} \leftarrow \frac{{{\pmb{V}^q}^\top \pmb{V}^r}}{{\|{\pmb{V}^q\|}\cdot\|\pmb{V}^r\|}} \nonumber 
        \end{align}
    \UNTIL{$s_{qr} > \tau$}
\STATE $\{\pmb{P}^q_{LR}, \pmb{P}^q_{HR}\} \leftarrow \{\pmb{P}^q_{LR}, \pmb{P}^q_{HR}\} \cup \{\pmb{p}^q_{LR}, \pmb{p}^q_{HR}\}$
\STATE $\{\pmb{P}^r_{LR}, \pmb{P}^r_{HR}\} \leftarrow \{\pmb{P}^r_{LR}, \pmb{P}^r_{HR}\} \cup \{\pmb{p}^r_{LR}, \pmb{p}^r_{HR}\}$
\ENDFOR
\end{algorithmic}
\label{alg:mini-batch}
\end{algorithm}

Our RefQSR was evaluated on the Urban100 dataset~\cite{huang2015single} as well as HR datasets, including the Test2K and Test4K datasets~\cite{kong2021classsr}. For patch-wise inference, each input image was cropped into patches with the size of 72$\times$72, 96$\times$96, and 96$\times$96 for Urban100, Test2K, and Test4K, respectively, with six overlapping boundary pixels between patches. We measured the peak signal-to-noise ratio (PSNR) and structural similarity index (SSIM)~\cite{wang2004image} between ground-truth HR and SR images as SR performance metrics. Furthermore, we used the average feature quantization ratio (FQR)~\cite{hong2022cadyq} and the number of operations weighted by the bitwidths of the operands (BitOPs)~\cite{van2020bitops} in the feature extraction stages of the SR networks to evaluate computational complexity, as outlined in~\cite{van2020bitops,hong2022cadyq}. The BitOPs were measured on images of sizes 1280$\times$720, 2048$\times$1080, and 4096$\times$2160 for Urban100, Test2k, and Test4k, respectively.

\begin{table*}[t]
\centering
\renewcommand{\tabcolsep}{1.3mm}
\caption{Quantitative comparisons of various quantization methods on SRResNet for a scale factor of 4. RefQSR is tested using various ($b_{high}$-$b_{low}$) combinations. $\delta$ represents a dynamically selected bit precision.}
\label{tab:exp-srresnet}
    \begin{tabular}{l cccc cccc cccc}
        \toprule
        \multirow{2}{*}{Model} & 
        \multicolumn{4}{c}{Urban100} & \multicolumn{4}{c}{Test2K} & \multicolumn{4}{c}{Test4K} \\
        \cmidrule(lr){2-5} \cmidrule(lr){6-9} \cmidrule(lr){10-13}& 
        BitOPs (G)$_\downarrow$ &FQR$_\downarrow$ & PSNR$_\uparrow$ & SSIM$_\uparrow$ &
        BitOPs (G)$_\downarrow$ &FQR$_\downarrow$ & PSNR$_\uparrow$ & SSIM$_\uparrow$ &
        BitOPs (G)$_\downarrow$ &FQR$_\downarrow$ & PSNR$_\uparrow$ & SSIM$_\uparrow$ \\
        \midrule
        SRResNet&148.32 &32.00&25.86&0.779&349.02 &32.00&27.53&0.771&1442.41 &32.00&28.91&0.818\\
        \midrule
        \midrule
        SRResNet-DDTB (8bit)&9.27 &8.00&25.89&0.778&21.81 &8.00&27.54&0.771&90.15 &8.00&28.93&0.819\\
        +RefQSR (8-5bit)&7.37 &6.81&\bf{25.98}&\bf{0.782}&16.57&6.60&\bf{27.57}&\bf{0.773}&65.49&6.40&\bf{28.95}&\bf{0.820}\\
        +RefQSR (8-4bit)&7.02&6.41&25.97&0.781&15.60&6.13&27.56&0.772&60.89&5.87&28.94&0.819\\
        +RefQSR (8-3bit)&\bf{6.67}&\bf{6.01}&25.95&0.781&\bf{14.63}&\bf{5.66}&27.55&0.772&\bf{56.31}&\bf{5.34}&28.93&0.819\\
        \midrule
        SRResNet-DDTB (4bit)&2.32&4.00&25.77&0.775&5.45&4.00&27.50&0.770&22.54&4.00&28.87&0.817\\
        +RefQSR (4-3bit)&2.14&3.60&\bf{25.92}&\bf{0.781}&4.95&3.53&\bf{27.54}&\bf{0.772}&20.23&3.47&\bf{28.92}&\bf{0.819}\\
        +RefQSR (4-2bit)&\bf{1.79}&\bf{3.20}&25.73&0.775&\bf{3.99}&\bf{3.07}&27.46&0.769&\bf{15.67}&\bf{2.93}&28.80&0.814\\
        \midrule
        \midrule
        SRResNet-PAMS (8bit)&9.27&8.00&25.87&0.778&21.81&8.00&27.53&0.771&90.15&8.00&28.91&0.818\\
        +RefQSR (8-5bit)&7.37 &6.81&\bf{25.96}&\bf{0.781}&16.57&6.60&\bf{27.56}&\bf{0.772}&65.49&6.40&\bf{28.95}&\bf{0.819}\\
        +RefQSR (8-4bit)&7.02&6.41&25.94&0.780&15.60&6.13&27.54&\bf{0.772}&60.89&5.87&28.93&\bf{0.819}\\
        +RefQSR (8-3bit)&\bf{6.67}&\bf{6.01}&25.88&0.778&\bf{14.63}&\bf{5.66}&27.51&0.770&\bf{56.31}&\bf{5.34}&28.89&0.817\\
        \midrule 
        SRResNet-PAMS (4bit)&2.32&4.00&25.64&0.771&5.45&4.00&27.46&0.768&22.54&4.00&28.81&0.815\\
        +RefQSR (4-3bit)&2.14&3.60&\bf{25.82}&\bf{0.777}&4.95&3.53&\bf{27.50}&\bf{0.771}&20.23&3.47&\bf{28.88}&\bf{0.818}\\
        +RefQSR (4-2bit)&\bf{1.79}&\bf{3.20}&25.50&0.765&\bf{3.99}&\bf{3.07}&27.38&0.765&\bf{15.67}&\bf{2.93}&28.71&0.813\\
        \midrule
        \midrule
        SRResNet-CADyQ ($\delta$bit)&6.02&5.18&25.89&\bf{0.780}&12.78&4.67&27.52&\bf{0.771}&51.45&4.55&28.90&\bf{0.818}\\
        +RefQSR ($\delta$-4bit)&4.28&4.04&\bf{25.92}&\bf{0.780}&9.81&4.00&\bf{27.54}&\bf{0.771}&39.92&4.00&\bf{28.92}&\bf{0.818}\\
        +RefQSR ($\delta$-3bit)&\bf{3.86}&\bf{3.64}&25.89&0.779&\bf{8.61}&\bf{3.51}&27.53&\bf{0.771}&\bf{34.67}&\bf{3.47}&28.90&\bf{0.818}\\
        \midrule
        \midrule
        SRResNet-CABM ($\delta$bit)&5.24&4.52&25.85&0.777&11.73&4.30&27.50&0.770&47.67&4.23&28.87&0.817\\
        +RefQSR ($\delta$-4bit)&4.23&4.00&\bf{25.90}&\bf{0.779}&9.79&4.00&\bf{27.54}&\bf{0.771}&39.84&4.00&\bf{28.92}&\bf{0.818}\\
        +RefQSR ($\delta$-3bit)&\bf{3.82}&\bf{3.61}&25.85&0.778&\bf{8.54}&\bf{3.49}&27.52&0.770&\bf{34.30}&\bf{3.44}&28.89&0.817\\
        \bottomrule
    \end{tabular}
\end{table*}

\begin{table}[htb]
\renewcommand{\tabcolsep}{.3mm}
\centering
\caption{Performance evaluation of several clustering methods on the Urban100 dataset}
\label{tab:Ablation_ClustBlock}
    \begin{tabular}{lcccc}
        \toprule
        & BitOPs (G)$_\downarrow$ & \begin{tabular}{l} Clustering\\BitOPs (G)$_\downarrow$ \end{tabular} &FQR$_\downarrow$ & PSNR$_\uparrow$ \\
        \midrule
        CADyQ ($\delta$bit) &6.02&-&5.18&25.89  \\
        \midrule
        + RefQSR ($\delta$-3bit) & & & & \\
        ~~~w/ SCAN &27.07&22.56&3.97&25.90  \\ 
        ~~~w/ SeCu &27.14&22.56&3.98&25.91  \\ 
        ~~~w/ ClustBlock (our) &3.86&0.03&3.64&25.89  \\
        \bottomrule
    \end{tabular}
\end{table}

\subsubsection{Parameter studies}

Since RefQSR replaces several layers with RefER blocks, we included the results obtained using different numbers of RefERs in Fig.~\ref{fig:hyperparameter}(a). The highest PSNR was obtained when two RefERs were used in SRResNet, which accounts for approximately 10\% of the number of feature extraction layers. We used a similar rate for the other two SR models, requiring one RefER for CARN and two RefERs for ELAN.

We applied thresholding using $\tau$ in the inference and training stages. Fig.~\ref{fig:hyperparameter}(b) shows the results on different values of $\tau$, indicating that $\tau = 0.5$ is a reasonable choice to filter out query patches that are not closely matched to their reference patches.

\begin{table*}[!h]
\centering
\renewcommand{\tabcolsep}{1.6mm}

\caption{
Quantitative comparisons of various quantization methods on CARN and ELAN for a scale factor of 4. RefQSR is tested using various ($b_{high}$-$b_{low}$) combinations.  $\delta$ represents a dynamically selected bit precision.}
\label{tab:exp-carn}
    \begin{tabular}{l cccc cccc cccc}
        \toprule
        \multirow{2}{*}{Model} & 
        \multicolumn{4}{c}{Urban100} & \multicolumn{4}{c}{Test2K} & \multicolumn{4}{c}{Test4K} \\
        \cmidrule(lr){2-5} \cmidrule(lr){6-9} \cmidrule(lr){10-13}& 
        BitOPs (G)$_\downarrow$ &FQR$_\downarrow$ & PSNR$_\uparrow$ & SSIM$_\uparrow$ &
        BitOPs (G)$_\downarrow$ &FQR$_\downarrow$ & PSNR$_\uparrow$ & SSIM$_\uparrow$ &
        BitOPs (G)$_\downarrow$ &FQR$_\downarrow$ & PSNR$_\uparrow$ & SSIM$_\uparrow$ \\
        \midrule
        CARN&83.43&32.00&26.07&0.784&196.33&32.00&27.58&0.773&811.35&32.00&28.95&0.820\\
        \midrule
        \midrule
        CARN-DDTB (8bit)&5.21&8.00&\bf{26.06}&\bf{0.784}&12.27&8.00&27.58&\bf{0.773}&50.71&8.00&28.96&\bf{0.820}\\
        +RefQSR (8-5bit)&4.20&6.85&\bf{26.06}&\bf{0.784}&9.48&6.65&\bf{27.59}&\bf{0.773}&37.59&6.47&\bf{28.98}&\bf{0.820}\\
        +RefQSR (8-4bit)&4.02&6.48&26.05&0.783&8.95&6.21&27.58&\bf{0.773}&35.12&5.96&28.97&\bf{0.820}\\
        +RefQSR (8-3bit)&\bf{3.83}&\bf{6.09}&26.04&\bf{0.784}&\bf{8.43}&\bf{5.76}&27.57&\bf{0.773}&\bf{32.65}&\bf{5.45}&28.96&\bf{0.820}
        \\
        \midrule
        CARN-DDTB (4bit)&1.30&4.00&25.89&0.779&3.07&4.00&27.52&0.770&12.68&4.00&28.89&0.818\\
        +RefQSR (4-3bit)&1.22&3.62&\bf{25.98}&\bf{0.782}&2.83&3.55&\bf{27.55}&\bf{0.772}&11.60&3.49&\bf{28.94}&\bf{0.819}\\
        +RefQSR (4-2bit)&\bf{1.03}&\bf{3.23}&25.93&0.780&\bf{2.31}&\bf{3.11}&27.53&\bf{0.772}&\bf{9.15}&\bf{2.98}&28.91&0.818\\
        \midrule
        \midrule
        CARN-PAMS (8bit)&5.21&8.00&\bf{26.05}&\bf{0.783}&12.27&8.00&27.57&\bf{0.773}&50.71&8.00&28.95&\bf{0.820}\\
        +RefQSR (8-5bit)&4.20&6.85&\bf{26.05}&\bf{0.783}&9.48&6.65&\bf{27.59}&\bf{0.773}&37.59&6.47&\bf{28.97}&\bf{0.820}\\
        +RefQSR (8-4bit)&4.02&6.48&26.04&\bf{0.783}&8.95&6.21&27.58&\bf{0.773}&35.12&5.96&28.96&0.819\\
        +RefQSR (8-3bit)&\bf{3.83}&\bf{6.09}&25.99&0.782&\bf{8.43}&\bf{5.76}&27.55&0.772&\bf{32.65}&\bf{5.45}&28.94&0.819\\
        \midrule 
        CARN-PAMS (4bit)&1.30&4.00&25.75&0.773&3.07&4.00&27.46&0.768&12.68&4.00&28.82&0.815\\
        +RefQSR (4-3bit)&1.22&3.62&\bf{25.84}&\bf{0.776}&2.83&3.55&\bf{27.49}&\bf{0.770}&11.60&3.49&\bf{28.86}&\bf{0.817}\\
        +RefQSR (4-2bit)&\bf{1.03}&\bf{3.23}&25.75&0.773&\bf{2.31}&\bf{3.11}&27.46&0.768&\bf{9.15}&\bf{2.98}&28.80&0.815\\
        \midrule
        \midrule
        CARN-CADyQ ($\delta$bit)&3.13&4.79&25.93&0.779&6.67&4.33&27.53&0.771&27.29&4.29&28.92&\bf{0.818}\\
        +RefQSR ($\delta$-4bit)&2.52&4.19&\bf{25.96}&\bf{0.781}&5.78&4.15&\bf{27.55}&\bf{0.772}&23.52&4.14&\bf{28.93}&\bf{0.818}\\
        +RefQSR ($\delta$-3bit)&\bf{2.20}&\bf{3.65}&25.93&0.780&\bf{4.90}&\bf{3.51}&27.53&0.771&\bf{19.78}&\bf{3.47}&28.89&\bf{0.818}\\
        \midrule
        \midrule
        CARN-CABM ($\delta$bit)&2.80&4.29&25.92&0.780&6.38&4.16&27.53&0.771&26.18&4.13&28.91&0.818\\
        +RefQSR ($\delta$-4bit)&2.44&4.07&\bf{25.96}&\bf{0.781}&5.61&4.04&\bf{27.54}&\bf{0.772}&22.86&4.03&\bf{28.92}&\bf{0.819}\\
        +RefQSR ($\delta$-3bit)&\bf{2.18}&\bf{3.63}&25.90&0.779&\bf{4.88}&\bf{3.51}&27.53&0.771&\bf{19.67}&\bf{3.46}&28.90&0.818\\
        \bottomrule
        \\
        \midrule
        ELAN&104.29&32.00&26.54&0.800&245.41&32.00&27.71&0.779&1014.20&32.00&29.14&0.825\\
        \midrule
        \midrule
        ELAN-DDTB (8bit)&10.32    &8.00     &\bf{26.48} &\bf{0.798} &     24.29&     8.00&\bf{27.71}&\bf{0.778}&100.36     &8.00     &\bf{29.13}&\bf{0.825}\\
        +RefQSR (8-5bit)&8.25     &6.72     &26.47      &0.797      &     18.85&     6.50&27.70     &\bf{0.778}&74.77     &6.29     &\bf{29.13}     &0.824\\
        +RefQSR (8-4bit)&7.85     &6.30     &26.35      &0.794      &     17.46&     6.00&27.68     &0.776     &68.22     &5.72     &29.09     &0.821\\
        +RefQSR (8-3bit)&\bf{7.36}&\bf{5.87}&26.19      &0.789      &\bf{16.10}&\bf{5.51}&27.62     &0.772     &\bf{61.81}&\bf{5.15}&29.02     &0.819\\
        \midrule 
        ELAN-DDTB (4bit)&3.85     &     4.00&25.88     &0.780     &9.05     &4.00     &27.54     &0.771&37.42     &4.00     &28.92&0.819\\
        +RefQSR (4-3bit)&3.47     &     3.57&\bf{25.99}&\bf{0.784}&8.02     &3.50     &\bf{27.57}&\bf{0.772}&32.58     &3.43     &\bf{28.93}     &\bf{0.820}\\
        +RefQSR (4-2bit)&\bf{2.99}&\bf{3.15}&25.74     &0.759     &\bf{6.68}&\bf{3.00}&27.45     &0.761     &\bf{26.29}&\bf{2.86}&28.70     &0.809\\
        \midrule
        \midrule
        ELAN-PAMS (8bit)&10.32    &8.00     &\bf{26.47} &\bf{0.798} &     24.29&     8.00&\bf{27.70}&\bf{0.778}&100.36     &8.00     &\bf{29.13}&\bf{0.823}\\
        +RefQSR (8-5bit)&8.25     &6.72     &26.42      &0.797      &     18.85&     6.50&27.69     &\bf{0.778}&74.77     &6.29     &29.11     &\bf{0.823}\\
        +RefQSR (8-4bit)&7.85     &6.30     &26.32      &0.794      &     17.46&     6.00&27.66     &0.776     &68.22     &5.72     &29.06     &0.821\\
        +RefQSR (8-3bit)&\bf{7.36}&\bf{5.87}&26.13      &0.789      &\bf{16.10}&\bf{5.51}&27.59     &0.772     &\bf{61.81}&\bf{5.15}&28.98     &0.819\\
        \midrule 
        ELAN-PAMS (4bit)&3.85     &     4.00&25.65     &0.773     &9.05     &4.00     &27.47     &\bf{0.769}&37.42     &4.00     &\bf{28.84}&\bf{0.816}\\
        +RefQSR (4-3bit)&3.47     &     3.57&\bf{25.68}&\bf{0.774}&8.02     &3.50     &\bf{27.49}&\bf{0.769}&32.58     &3.43     &28.83     &0.815\\
        +RefQSR (4-2bit)&\bf{2.99}&\bf{3.15}&25.30     &0.759     &\bf{6.68}&\bf{3.00}&27.32     &0.761     &\bf{26.29}&\bf{2.86}&28.61     &0.809\\
        \midrule
        \midrule
        ELAN-CADyQ ($\delta$bit)&6.65    &5.10     &\bf{26.16}&\bf{0.789}&14.30    &4.74     &\bf{27.60}&\bf{0.774}&57.58     &4.62&\bf{29.03}&\bf{0.821}\\
        +RefQSR ($\delta$-4bit)&4.86     &4.20     &26.15     &\bf{0.789}&10.88    &4.07     &\bf{27.60}&\bf{0.774}&43.67     &4.02&\bf{29.03}&\bf{0.821}\\
        +RefQSR ($\delta$-3bit)&\bf{4.38}&\bf{3.79}&25.97     &0.785     &\bf{9.56}&\bf{3.58}&27.55     &0.772     &\bf{37.68}&\bf{3.47}&28.92&0.818\\
        \midrule
        \midrule
        ELAN-CABM ($\delta$bit)&5.73&4.56&\bf{26.13}&\bf{0.788}&13.05&4.42&\bf{27.60}&\bf{0.774}&53.04&4.35&\bf{29.02}&\bf{0.821}\\
        +RefQSR ($\delta$-4bit)&4.78&4.19&26.12&\bf{0.788}&10.74&4.07&\bf{27.60}&\bf{0.774}&43.18&4.02&29.01&0.820\\
        +RefQSR ($\delta$-3bit)&\bf{4.29}&\bf{3.77}&25.95&0.782&\bf{9.37}&\bf{3.56}&27.53&0.770&\bf{37.19}&\bf{3.47}&28.91&0.818\\
        \bottomrule
    \end{tabular}
\end{table*}

\begin{figure*}[tb]
\centering
\includegraphics[width=1.\linewidth]{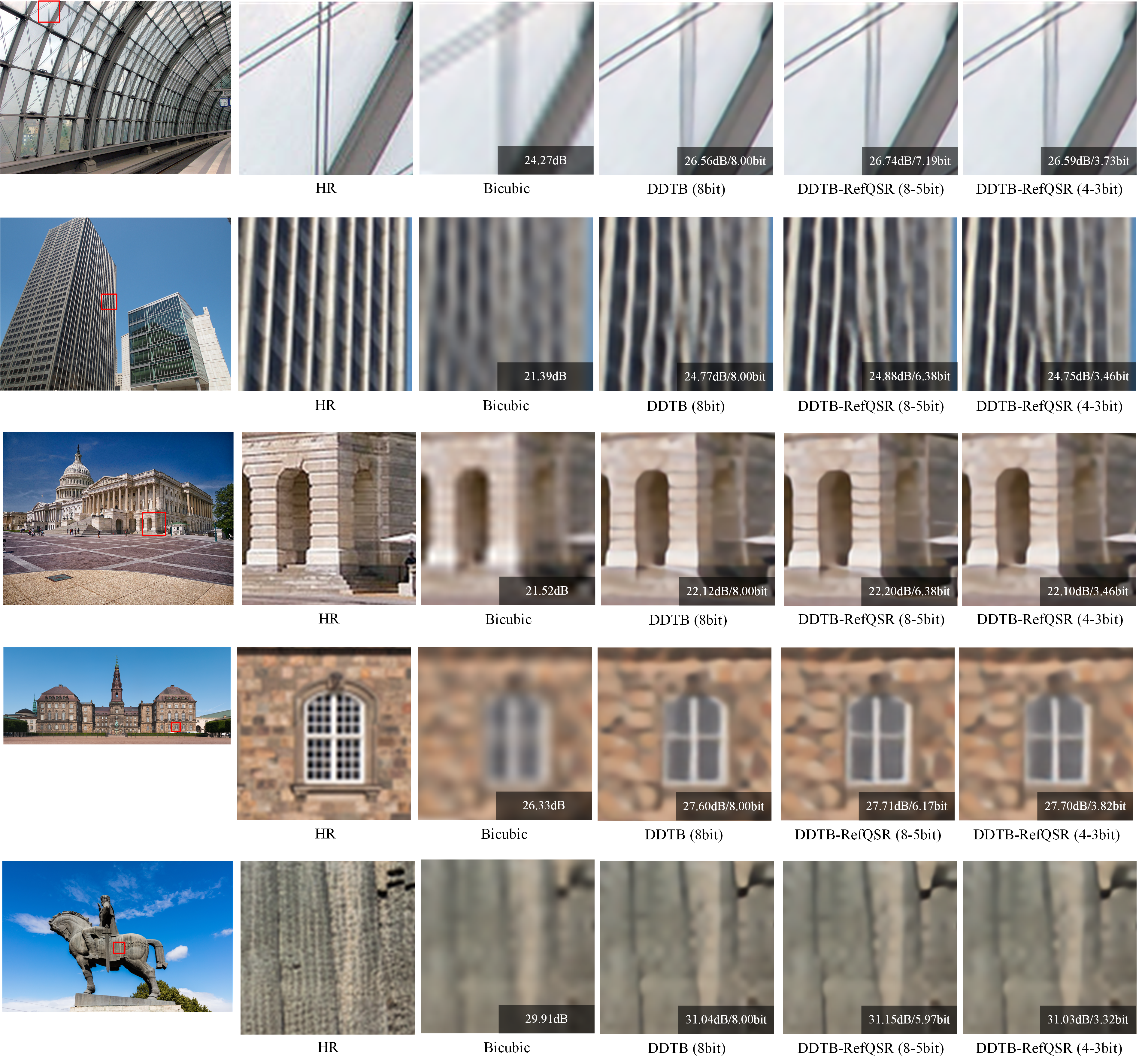}
\caption{Visual comparison of DDTB and DDTB-RefQSR on SRResNet.}
\label{fig:result-1}
\end{figure*}

\begin{figure*}[htb]
\centering
\includegraphics[width=1.\linewidth]{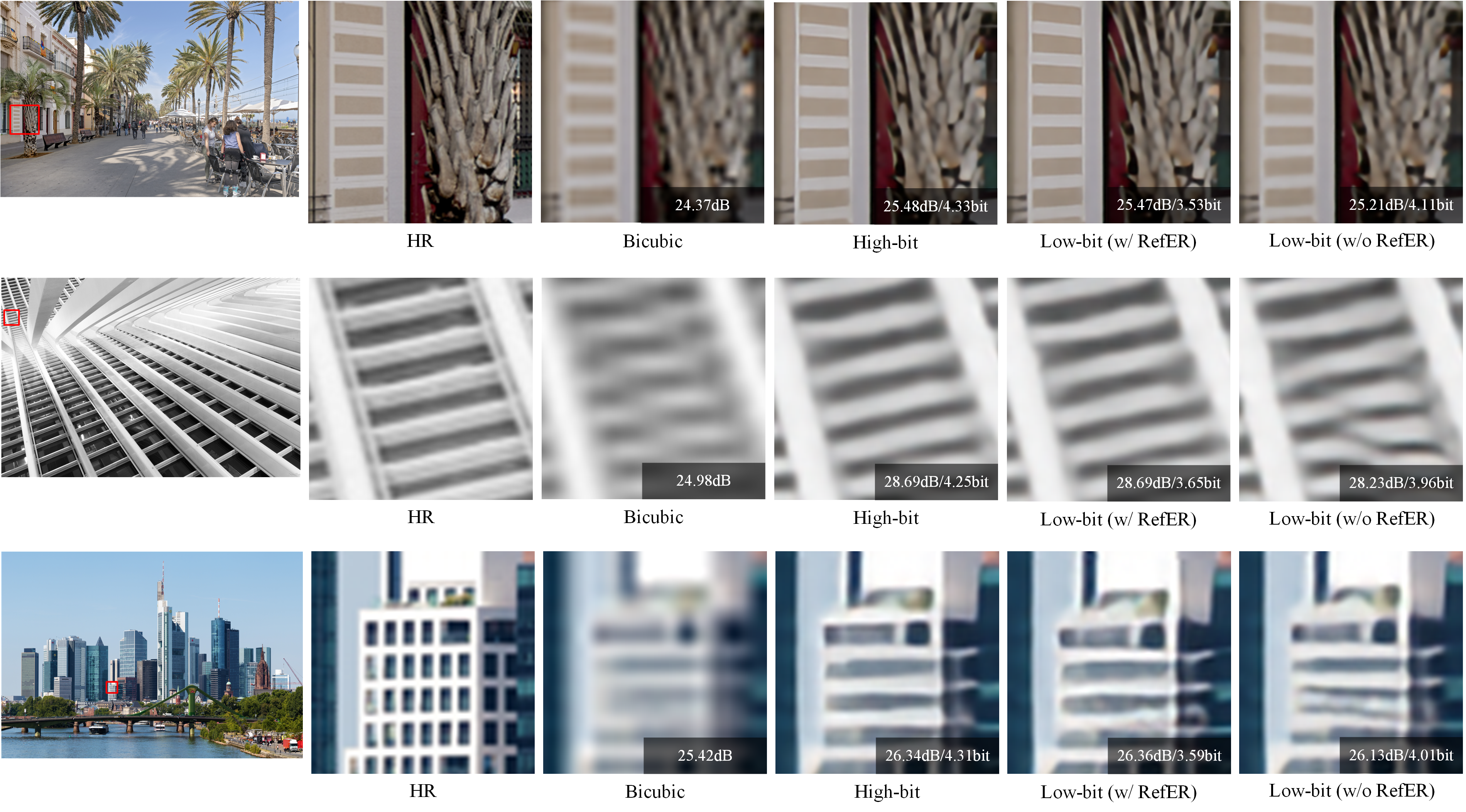}
\caption{Visual comparison of CADyQ-RefQSR ($\delta$-3bit) under different settings. The same patches (red boxes in the first column) are processed as reference patches using $b_{high}$ = $\delta$ (the fourth column) and as query patches using RefER with $b_{low}$ = 3 (the fifth column), and without using RefER with $b_{low}$ = 3 (the sixth column).}
\label{fig:result-ab}
\end{figure*}

\subsection{Quantitative Results}
First, to verify the effectiveness of the proposed ClustBlock, we compared ours
with two recent methods, SCAN~\cite{van2020scan} and SeCu~\cite{qian2023secu}, using SRResNet-CADyQ. As Table~\ref{tab:Ablation_ClustBlock} shows, ClustBlock performed comparably with these recent methods in PSNRs while requiring significantly fewer BitOPs. Thus, ClustBlock can be considered a cost-effective solution for RefQSR.

To demonstrate the effectiveness and applicability of RefQSR, we tested RefQSR on fixed-precision quantization methods: PAMS and DDTB, and mixed-precision methods: CADyQ and CABM. Tables~\ref{tab:exp-srresnet} and \ref{tab:exp-carn} provide the comparison results on SRResNet, CARN, and ELAN. RefQSR was evaluated with various combinations of high- and low-bit precision settings.

As shown in Table~\ref{tab:exp-srresnet}, RefQSR led to significant reductions in computational complexity of the baseline quantization methods on SRResNet. In particular, the performance improvement is noticeable in the fixed-precision methods. For example, DDTB-RefQSR (4-3bit) achieved a reduction of 76.9\% in BitOPs while maintaining a similar PSNR score compared to DDTB (8bit) in Urban100. The performance improvement on the mixed-precision methods is also non-negligible. For example, CADyQ-RefQSR ($\delta$-3bit) was able to reduce BitOPs by 35.9\% while still maintaining a PSNR score similar to that of CADyQ in Urban100. Note that $\delta$ represents an adaptive bit precision determined based on the complexity of the input images~\cite{hong2022cadyq, tian2023cabm}.

In addition, the use of RefQSR resulted in consistent improvements for CARN, as demonstrated in Table~\ref{tab:exp-carn}. For example, for CARN, DDTB-RefQSR (4-2bit) achieved a reduction of 20.8\% in BitOPs while exhibiting a slightly higher PSNR score compared to DDTB (4bit) in Urban100, and CADyQ-RefQSR ($\delta$-3bit) enabled a reduction of 29.7\% in BitOPs while showing a similar PSNR score compared to CADyQ in Urban100. The results on a recent SR network, ELAN, show trends similar to those obtained using SRResNet and CARN, \textit{i.e.}, RefQSR produces images with similar quality while consuming significantly lower BitOPs and FQRs. For example, CADyQ-RefQSR ($\delta$-3bit) reduced BitOPs in Urban100 by 34.1\% compared to CADyQ. More results with different upsampling factors can be found on the project page\footnote{\url{https://jimmy9704.github.io/RefQSR/}}. 

\subsection{Qualitative Results}
Fig.~\ref{fig:result-1} provides several SR results for visual comparisons. On the one hand, DDTB-RefQSR (8-5bit) produced SR images with sharp edges and rich texture regions compared to DDTB (8bit) while consuming fewer BitOPs. On the other hand, DDTB-RefQSR (4-3bit) produced SR images comparable to those of DDTB (8bit) while using significantly fewer BitOPs. More results for visual comparisons with other baseline quantization methods can be found on the project page. 

Fig.~\ref{fig:result-ab} further provides the SR results obtained using different configurations of RefQSR. First, given the trained CADyQ-RefQSR, we treated all patches as reference patches for the test images and processed them using dynamically selected bit precision, \textit{i.e.}, $b_{high}$ = $\delta$. Second, we trained CADyQ-RefQSR while replacing RefER blocks with high-bit ResBlocks. During the inference stage, we clustered patches using ClustBlock and processed reference patches using $b_{high}$ = $\delta$. The query patches were processed using $b_{low}$ = 3 except for the last two ResBlocks with $b_{high}$ = $\delta$. Compared to the first method, the proposed method produced images of comparable quality while consuming significantly fewer bits. Compared to the second method, the proposed method produced images with significantly higher quality while consuming slightly fewer bits. These results support the effectiveness of the proposed reference-based quantization method.

\subsection{Ablation Studies}

Table~\ref{tab:Ablation_RefER} shows the results on Urban100 with various modifications of RefER using SRResNet-CADyQ as a baseline. When the RefER block was not used but replaced by high-bit ResBlocks (w/o RefER), FQR and BitOPs were reduced at the cost of decreased SR performance. Next, we fed the query patch as the reference patch in RefER (self-reference) and obtained performance degradation in PSNR and SSIM. When RefER was used without applying pre-processing in (\ref{eqn: preprocess}) (w/o pre-process), we achieved considerable improvements in FQR at the cost of significantly increased BitOPs. Furthermore, the concatenation of the query feature and the aligned reference feature (w/ concat) was found to be less effective in BitOPs than taking a difference of them as RefER. These results indicate that our RefER is designed to effectively restore quantization errors with minimal computational overhead.

\begin{table}[t]
\centering
\caption{Ablation study on several RefER variants on the Urban100 dataset}
\label{tab:Ablation_RefER}
    \begin{tabular}{lcccc}
        \toprule
        & BitOPs (G)$_\downarrow$ &FQR$_\downarrow$ & PSNR$_\uparrow$ & SSIM$_\uparrow$ \\
        \midrule
        CADyQ ($\delta$bit) &6.02&5.18&25.89&0.780  \\
        \midrule
        w/o RefER &4.32&4.10&25.79&0.775  \\ 
        Self-reference &4.08&3.85&25.78&0.775  \\ 
        w/o pre-process &6.29&3.74&25.86&0.778  \\ 
        w/ concat &4.18&3.79&25.90&0.779  \\ 
        \midrule
        +RefQSR ($\delta$-3bit) &3.86&3.64&25.89&0.779  \\
        \bottomrule
    \end{tabular}
\end{table}

\begin{table*}[!t]
\centering
\renewcommand{\tabcolsep}{4.5mm}
\caption{Complexity analysis of the quantized ResBlocks in RefQSR, measured for $\times$4 SR to obtain an image with 4k resolution}
\label{tab:complexity analysis}
    \begin{tabular}{ccccccc}
        \toprule
        \multicolumn{2}{c}{Model} & Params (M)$_\downarrow$ &BitOPs (G)$_\downarrow$ &FQR$_\downarrow$ &Latency (ms)$_\downarrow$ &Peak memory (MB)$_\downarrow$ \\
        \midrule
        \multicolumn{2}{c}{SRResNet-DDTB (8bit)} &0.30&90.15 &8.00 &33.11 &2.99\\
        \multicolumn{2}{c}{SRResNet-DDTB (4bit)} &0.15&22.54 &4.00 &26.21 &2.67\\
        \midrule
        \multirow{2.7}{*}{+RefQSR (8-4bit) }       & RefER($\times$2)   &0.04&4.20 &8.00&5.23  &3.54\\
                                                 \cmidrule(l){2-7}
                                                 & Total &0.34&55.03&5.87&29.53 &3.58\\
        \bottomrule
    \end{tabular}
\end{table*}

\begin{table}[!h]
\centering
\renewcommand{\arraystretch}{1.1} 
\caption{Comparison of image-wise and patch-wise inference on the Test4k dataset}
\label{tab:patch complexity}
    \begin{tabular}{lccc}
        \toprule
         Model&Patch size& BitOPs (G)$_\downarrow$ & {PSNR$_\uparrow$}  \\
        \midrule
        \multirow{2}{*}{SRResNet (32bit)} & Full image &1304.60&28.91  \\
         & 96 $\times$ 96 &1442.41 &28.91  \\
        \midrule
        \multirow{2}{*}{SRResNet-DDTB (8bit)} & Full image &81.54&28.93 \\
         & 96 $\times$ 96 &90.15&28.93  \\
        \midrule
        \multirow{5}{*}{~+RefQSR (8-3bit)} & 48 $\times$ 48 &67.91&28.92  \\
         & 72 $\times$ 72 &58.96&28.92  \\
         & 96 $\times$ 96 &56.31&28.93  \\
         & 256 $\times$ 256 &58.94&28.93 \\
         & 384 $\times$ 384 &66.83&28.93 \\
        \bottomrule
    \end{tabular}
\end{table}

\subsection{Complexity Analysis}\label{sec:complexity}
\subsubsection{RefQSR}

Table~\ref{tab:complexity analysis} presents the complexity analysis of the quantized ResBlocks of the SR networks with and without RefQSR. We measured the number of parameters~\cite{zhong2022dynamic}, BitOPs~\cite{van2020bitops}, FQR~\cite{hong2022cadyq}, latency, and peak memory required to obtain a 4k resolution image by $\times$4 SR as complexity metrics. In particular, latency was measured using the TVM toolkit~\cite{tvm}, as performed in CADyQ~\cite{hong2022cadyq}. For this measurement, the A100 GPU was used, which supports accelerations only for 8-bit and 4-bit. Thus, we measured the latency using $b_{high} = 8$, $b_{medium} = 4$, and $b_{low} = 4$. 

In terms of BitOPs, SRResNet-DDTB with RefQSR requires two additional RefER blocks. Despite the increase in BitOPs by additional blocks, the total BitOPs decreased by 39.0\% compared to SRResNet-DDTB (8-bit) due to low-bit precision processing of query patches. In terms of latency, SRResNet-DDTB with RefQSR required 29.53 ms for processing, where SRResNet-DDTB (8-bit) and SRResNet-DDTB (4-bit) required 33.11 ms and 26.21 ms, respectively.
Since the parameters are shared between high- and low-bit precision ResBlocks, RefQSR increased only 0.04M parameters. This slight increase in parameters is acceptable considering the significant reduction in BitOPs achieved by RefQSR. Last, peak memory also increased due to the additional blocks and the storage for the reference features. However, due to the patch-wise processing nature of RefQSR, the increase in the peak memory over SRResNet-DDTB (8-bit) is only 0.59 MB.

\subsubsection{Patch-wise inference}

Our RefQSR is only applicable in the patch-wise inference scenario, whereas several methods, such as DDTB, can be applied to both image-wise and patch-wise inference. Here, we conducted a comparative analysis of DDTB and DDTB-RefQSR using different patch sizes on the Test4K dataset. Table~\ref{tab:patch complexity} shows that SRResNet and SRResNet-DDTB increased 10.6\% and 10.6\% of BitOPs, respectively, due to patch-wise inference. However, with a proper choice of the patch size, \textit{i.e.}, 96$\times$96, our RefQSR overcame such an overhead of patch-wise inference by assigning fewer bits to query patches, resulting in much fewer BitOPs with similar PSNR scores. 

\section{Conclusion}
In this paper, we proposed RefQSR, a novel reference-based quantization method for image super-resolution, to address the high computational costs of SISR networks. RefQSR is the first method that explicitly uses image self-similarity for SISR network quantization. Specifically, RefQSR applies high-bit quantization to several representative patches and uses them as references for low-bit quantization of the remaining patches in an image. To this end, we developed dedicated patch clustering and reference-based error refinement modules and integrated them into existing SISR networks. Experimental results demonstrate the effectiveness of RefQSR in various SISR networks and quantization methods. RefQSR achieves high SISR performance by leveraging image self-similarity while significantly reducing computational costs, making it a promising solution for SISR network quantization. 

Future work is being considered that addresses several limitations of RefQSR. First, ClustBlock is pre-trained separately. We will explore the potential of end-to-end learning of patch clustering and SR to exploit image self-similarity more effectively. Second, the naive implementation of RefQSR can be less friendly to parallel processing. We will optimize RefQSR for parallel processing and test RefQSR on different hardware platforms that support mixed-precision operations.
Finally, our idea of reference-based quantization is not restricted to SR. We will verify the effectiveness of the proposed methodology on other low-level vision tasks, such as image deblurring and denoising. 

\bibliographystyle{IEEEtran}
\bibliography{cite}
 
\vspace{11pt}


\vspace{11pt}


\vfill

\end{document}